\documentclass{article}
\usepackage{arxiv}
\usepackage{times}
\usepackage{epsfig}
\usepackage{graphicx}
\graphicspath{ {./images/} }
\usepackage{amsmath}
\usepackage{amssymb}
\usepackage{subfig}
\usepackage{caption}
\usepackage{mathtools}

\usepackage{array}
\usepackage{tabularx}
\usepackage{booktabs}
\usepackage{multirow}
\usepackage{textcomp}
\usepackage[accsupp]{axessibility}
\usepackage{url}
\usepackage{fullpage}
\usepackage{times}
\usepackage{fancyhdr}
\usepackage[dvipsnames]{xcolor}
\usepackage[ruled,vlined]{algorithm2e}
\include{pythonlisting}

\usepackage[T1]{fontenc}
\usepackage{enumitem}

\usepackage[utf8]{inputenc} 
\usepackage[T1]{fontenc}    
\usepackage{hyperref}       
\usepackage{url}            
\usepackage{booktabs}       
\usepackage{amsfonts}       
\usepackage{nicefrac}       
\usepackage{microtype}      
\usepackage{lipsum}		
\usepackage{graphicx}
\usepackage{natbib}
\usepackage{doi}

\title{Moving Object Detection for Event-based Vision using Graph Spectral Clustering}

\author{Anindya Mondal$^1$\thanks{Authors have equal contributions.} , Shashant R$^1$\footnotemark[1] , Jhony H. Giraldo$^2$, Thierry Bouwmans$^2$, Ananda S. Chowdhury$^1$ \\
$^1$ Department of Electronics and Telecommunication Engineering, Jadavpur University, India\\
$^2$ 
Laboratoire Mathématiques, Image et Applications (MIA), La Rochelle Université, France \\
{\tt\small \{anindyam.jan, shashant7699\}@gmail.com, \{jgiral01, tbouwman\}@univ-lr.fr,} \\
{\tt\small as.chowdhury@jadavpuruniversity.in}}

\date{}

\renewcommand{\headeright}{ICCVW 2021}
\renewcommand{\undertitle}{Accepted in IEEE/CVF ICCVW 2021 (Virtual)}

\hypersetup{
pdftitle={A template for the arxiv style},
pdfsubject={q-bio.NC, q-bio.QM},
pdfauthor={David S.~Hippocampus, Elias D.~Striatum},
pdfkeywords={First keyword, Second keyword, More},
}

\begin{document}
\maketitle

\begin{abstract}
	Moving object detection has been a central topic of discussion in computer vision for its wide range of applications like in self-driving cars, video surveillance, security, and enforcement. Neuromorphic Vision Sensors (NVS) are bio-inspired sensors that mimic the working of the human eye. Unlike conventional frame-based cameras, these sensors capture a stream of asynchronous ‘events’ that pose multiple advantages over the former, like high dynamic range, low latency, low power consumption, and reduced motion blur. However, these advantages come at a high cost, as the event camera data typically contains more noise and has low resolution. Moreover, as event-based cameras can only capture the relative changes in brightness of a scene, event data do not contain usual visual information (like texture and color) as available in video data from normal cameras. So, moving object detection in event-based cameras becomes an extremely challenging task. In this paper, we present an unsupervised Graph Spectral Clustering technique for Moving Object Detection in Event-based data (GSCEventMOD). We additionally show how the optimum number of moving objects can be automatically determined. Experimental comparisons on publicly available datasets show that the proposed GSCEventMOD algorithm outperforms a number of state-of-the-art techniques by a maximum margin of 30$\%$. 
\end{abstract}


\section{Introduction}
Recent developments in materials engineering, fabrication technology, VLSI (very-large-scale-integration) design techniques, and neuro-science have facilitated the newly discovered concept of bio-inspired visual sensors and processors (\cite{chen2020event}). Event-based cameras are such neuromorphic bio-inspired visual sensors that mimic the mode of action of a biological retina, bringing a paradigm shift in computer vision technology (\cite{chen2020event}). One of the earliest contributions to put the concept of event-based cameras in action was made by Lichtsteiner \textit{et al.} (\cite{lichtsteiner2006128}), where they  developed an event-based neuromorphic sensor based on biological principles. These bio-inspired cameras capture a stream of asynchronous events in contrast to the traditional RGB cameras which acquire images at a fixed-frame rate as specified by an external clock (\cite{gallego2020event}). In event cameras, each pixel memorizes the log intensity each time an event is sent and incessantly monitors for a sufficient change in magnitude from this memorized value. When this change crosses a threshold value, an event is recorded by the camera, which is then transmitted by the sensor in form of its location $\{x,y\}$, its timestamp $t$ (in microseconds), and its polarity $p$ (\textit{i.e.}, whether the pixel had become brighter or darker). In Fig. \ref{fig:evcam}, we illustrate how an event camera works following (\cite{gallego2020event}). In event-based cameras,  sensors capture the per-pixel brightness changes (called events) asynchronously instead of measuring the absolute brightness of all pixels at a constant rate. As a result, traditional vision algorithms designed to process video data from frame-based cameras often become unsuitable for processing event data. 

\begin{figure*}
\begin{center}
\includegraphics[width=0.88\textwidth]{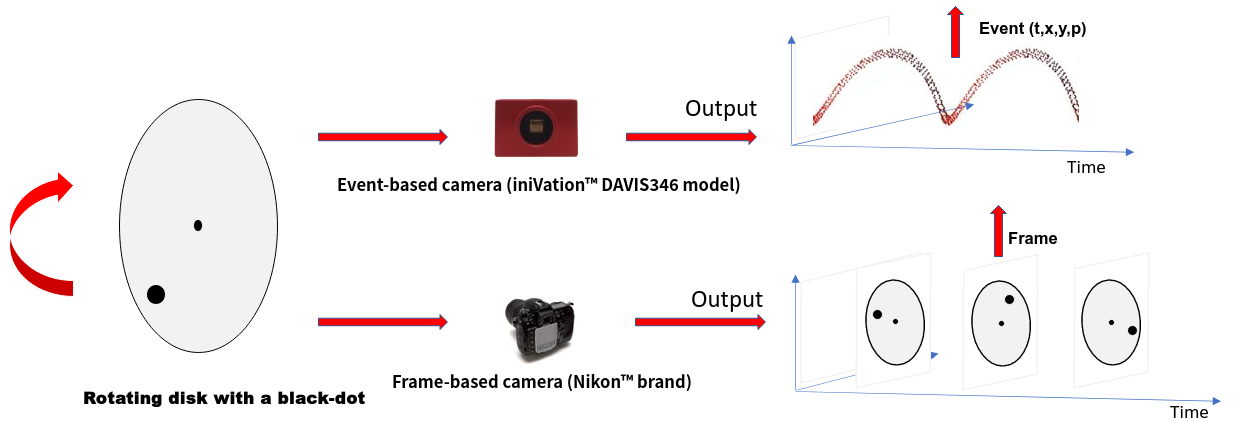}
\end{center}
   \caption{Visualization of the output from a neuromorphic vision sensor and a standard frame-based camera when facing a rotating disk with a black dot. Inspired by \cite{gallego2020event}}.
    \label{fig:evcam}
\label{fig:evcam}
\end{figure*}

Moving object detection is an important task in the field of computer vision which appears in autonomous vehicles, surveillance systems and alike. In moving object detection, we identify the physical movement of an object in a given region or area (\cite{kulchandani2015moving}). However, event-based moving object detection systems are far inferior to their frame-based counterparts (\cite{piccardi2004background,agarwal2016review}), because only a limited amount of reliable labels are available for training (\cite{bi2020graph}). In fact, this limitation restricts the use of deep learning in solving motion detection problems (from event data).   

Moving object detection in event-based cameras becomes further challenging as event cameras can only capture the relative changes in brightness of a scene and not usual visual features like color and texture (\cite{pikatkowska2012spatiotemporal}). 

Recently few attempts have been made to extend the task of moving object detection in neuromorphic vision, either by using parametric models (\cite{mitrokhin2018event,stoffregen2019event}) or by using traditional clustering algorithms (\cite{pikatkowska2012spatiotemporal,hinz2017online,chen2018neuromorphic}). Unfortunately, those parametric models are complex and require several assumptions to hold for their proper working. Furthermore, traditional clustering algorithms are sensitive to noise generated from the motion of the objects and sensor defects (temporal and shot noise) (\cite{chen2020event}).
Widespread use of graph-based representations have been noticed in recent computer vision and machine learning applications (\cite{giraldo2020graph,giraldo2021graphbgs,xia2021graph}). In some cases (like financial and banking data, social networks, mobility and traffic patterns, marketing preferences, fads, etc.), the data resides on irregular and complex structures, which can be efficiently tackled with graph-based methods (\cite{ortega2018graph}). Note that neuromorphic-based sensors activate asynchronously in time. So, the data streams are produced at irregular space-time coordinates which depends upon the scene activity (\cite{bi2020graph}). Therefore, by representing the events as graphs, one can maintain the event asynchronicity and sparsity and exploit their advantages (\cite{bi2020graph}).

Our method improves the previous works (\cite{pikatkowska2012spatiotemporal,chen2018neuromorphic,hinz2017online}) in multiple ways. Firstly, we do not need any prior knowledge about the actual number of moving objects in a scene. Secondly, our model is successful in detecting the moving objects from the noisy event data. We construct a similarity graph using k-Nearest Neighbors (k-NN). Then, graph spectral clustering is applied for detecting the moving objects. Our contributions can be summarized as follows:

\begin{itemize}[leftmargin=*]
    \itemsep0em
    \item We introduce graph-spectral clustering (\cite{martin2018robust,luo2003spectral,von2007tutorial,ng2001spectral,panda2017nystrom}) for detecting moving objects in event data. We use this method because it can handle clusters of arbitrary (including non-convex) shapes and it does not make any prior assumptions about the cluster shapes (\cite{meila2016spectral}).
    \item We show, using silhouette analysis (\cite{shutaywi2021silhouette}), how the actual number of moving objects in the event-based data can be automatically determined.
    \item We show that our method (GSCEventMOD) allows successful detection of moving objects in a scene. Experimental results show that GSCEventMOD performs better than previous state-of-the-art techniques on a publicly available dataset (\cite{almatrafi2020distance}). We also demonstrate the versatility of our method by testing it on a synthetically generated dataset.
\end{itemize}
The rest of the paper is organized as follows:
In Section 2 we briefly discuss the related works. In Section 3, we describe our proposed method. The details of our experiments and comparisons with the state-of-the-art are presented in Section 4. Finally, in Section 5 we conclude the paper with outlines for direction of future work.

\begin{figure*}
\begin{center}
\includegraphics[width=\textwidth]{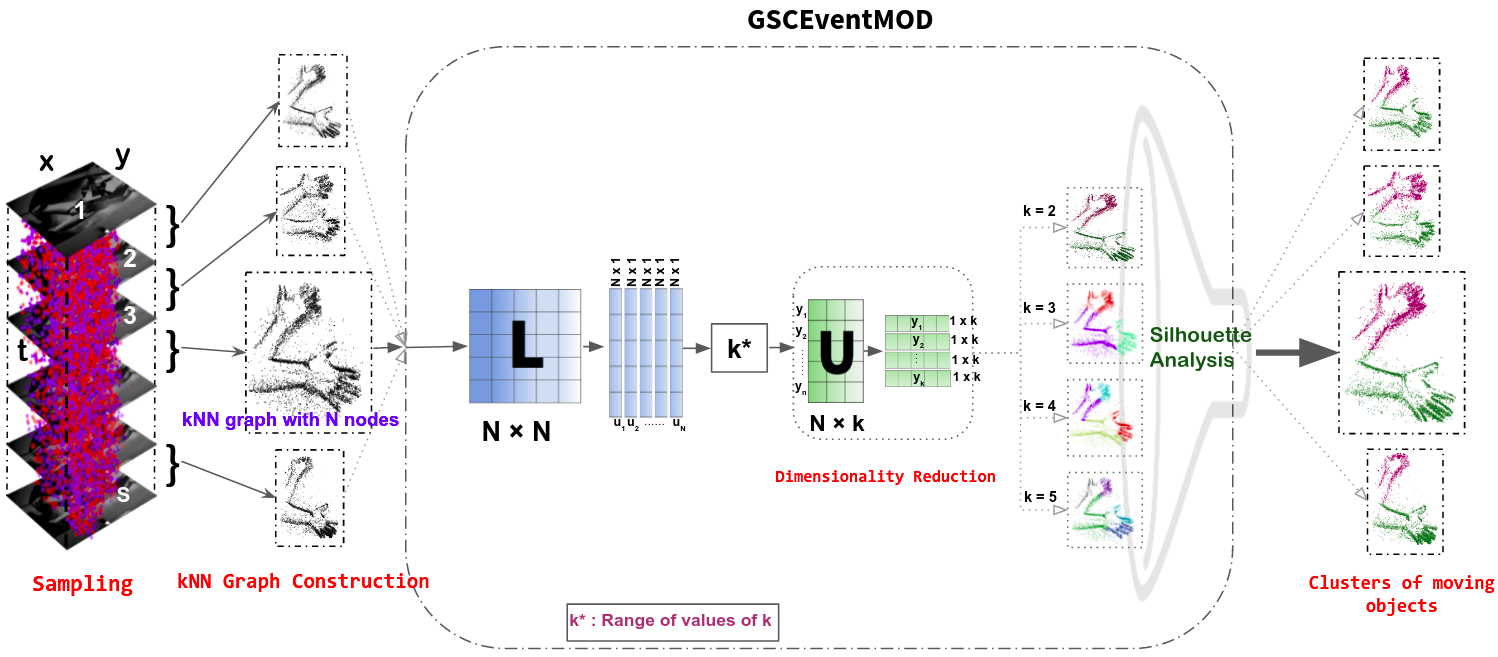}
\end{center}
   \caption{An illustrative overview of the proposed GSCEventMOD. First, we sample the event space-time volume on the basis of the timestamps of the corresponding grayscale images. Secondly, we construct a similarity graph using k-NN. We also perform eigendecomposition on the Graph Laplacian and take the first $k$ eigenvectors to get the moving objects as clusters. We use silhouette analysis for determining the optimal value of $k$.}
    \label{fig:pipeline}
\label{fig:short}
\end{figure*}

\section{Related work}
In this section, we briefly discuss the available techniques for moving object detection in frame-based cameras and see how they are extended to neuromorphic vision. Many classical approaches that are proposed for moving object detection are based on the geometrical understanding of the scene like in (\cite{menze2015object}). There are several deep learning-based approaches too, like in (\cite{huang2019deep,zhu2020moving}), where they have used multi-layered convolutional neural networks for detecting moving objects. However, using deep learning has its own disadvantages, for example, the models are very complex and they require a large amount of labeled data to avoid over-fitting. Indeed, there are no general answers in the literature about the sample complexity required in the deep learning regimen (\cite{giraldo2020graph}).

Also during high-speed motion, traditional RGB cameras suffer from high motion blur and perform badly in challenging situations (like abrupt variation in ambient brightness, remote areas with power scarcity, etc.). Here, event-based cameras (\cite{chen2020event,gallego2020event}) can be useful (\cite{rebecq2019high}). This is because these neuromorphic sensors capture the scene dynamics only (as the change in brightness occurs only when there is motion in the scene) and detection of the moving objects from the static background is done by the sensor itself (\cite{pikatkowska2012spatiotemporal}). So, the task of moving object detection may seem trivial if there is only one object in a scene as all the generated events correspond to the motion of the object (ignoring noise). However, when there are multiple moving objects in a scene, the task of detecting them all becomes really difficult. This is because unlike the frame-based cameras, neuromorphic vision sensors capture only the binary changes in brightness, which do not contain much visual information (\cite{pikatkowska2012spatiotemporal}).

In 2012, Piątkowska \textit{et al.} (\cite{pikatkowska2012spatiotemporal}) have studied the possibility of using Gaussian Mixture Models (GMMs) (\cite{reynolds2009gaussian}) for multiple persons tracking using event-based cameras. However, a model using GMM is too sensitive to noise, as it assumes that each data point (here the events) is independent of its neighbors, thus ignoring the similarity relations among those points (\cite{nguyen2011gaussian}).
 
In 2018, Chen \textit{et al.} (\cite{chen2018neuromorphic}) and Hinz \textit{et al.} (\cite{hinz2017online}) have performed a preliminary multi-vehicle detection and tracking using classical clustering approaches (like DBSCAN (\cite{khan2014dbscan}), MeanShift (\cite{derpanis2005mean}), etc.).
However, these methods also perform poorly as they are sensitive to noise and require tuning of quite a few parameters (\cite{feng2018robust}). Here graph spectral clustering performs better, because it requires the tuning of a single parameter (i.e. the number of clusters). In this work, we show how the number of clusters can be determined automatically using silhouette analysis.

\section{Proposed Method}
In our work, we use graph-spectral clustering (\cite{luo2003spectral,martin2018robust}) for the task of detecting moving objects in event-based data. We show that the application of graph spectral clustering can find meaningful clusters of arbitrary shapes under realistic separations. The schematic of the proposed unsupervised model, termed as, Graph Spectral Clustering technique in Event data for Moving Object Detection (GSCEventMOD) is shown in Fig. \ref{fig:pipeline}. A sampling strategy is first employed for obtaining a small set of events to facilitate computationally efficient processing. The sampled events are then used to construct a k-NN graph. Then we apply spectral clustering on the k-NN graph. The clusters represent the moving objects.

\subsection{Sampling and Graph Construction}
The events generated due to the moving objects are treated as sparse point-cloud data in the 3-dimensional space-time volume (\cite{chen2020event}). Let us consider there are $M$ event points and corresponding $S$ grayscale images captured at uniform timestamps $T_i$ ($i$ ranging from 1 to $S$). To generate meaningful samples, we divide the whole space-time event volume into $S$ partitions. Each of these individual partitions consists of $P_i$ events that occurred before timestamp $T_i$. Thus, $\sum_{i=1}^{S} P_i = M$. 

We adopt a uniform sampling strategy for obtaining a small set of neuromorphic events. Let $N$ events are selected from each of these partitions. Each event-point in $N$ is represented as a tuple sequence:
\begin{equation} \label{tuple}
\{e_i\}_N = \{x_i , y_i , t_i\}_N,
\end{equation}
where $\{x_i, y_i\}$ indicates the spatial address at which the spike event had occurred, $t_i$ is the timestamp indicating when the event was generated and $N$ represents the total number of events. 

As the events are sparse in the spatio-temporal domain (image plane evolving in time), the underlying graph is generally unstructured (as opposed to the graph of pixels in an image, which is regular) (\cite{zhou2020event}). To address the irregularity, a graph is constructed with $N$ events using the popular k-NN strategy (\cite{ortega2018graph}). The neighborhood is based on the spatio-temporal similarity between the event points in the point cloud. Let us define a graph $G = G(V,E)$, where  $V$ is the set of nodes or vertices and $E$ is the set of edges. Here we represent each event $e_i = e_i(x_i , y_i , t_i)$ as a node $v_i$ in the graph ($v_i \in V$). We connect $v_i$ and $v_j$ with an edge $\epsilon_{ij}$ ($\epsilon_{ij} \in E$) if either $v_i$ is among the $k$ nearest neighbors of $v_j$ or $v_j$ is among the $k$ nearest neighbors of $v_i$. 

The value of $k$ in k-NN has been chosen carefully. A small value of $k$ would make the result sensitive to noise whereas a large value of $k$ would make the process computationally expensive. As there is no definite statistical method to find an optimal $k$ and there is abundance of noise in the event data (due to current limitations of neuromorphic vision sensors (\cite{chen2018neuromorphic}), we have experimentally set its value for each sequence. 

\subsection{Graph Laplacian}
Let $\mathbf{A}$ be the adjacency matrix of the graph $G$ with $\mathbf{A} = (a_{ij})_{i,j = 1,2,...,N}$, where the set of vertices is $V = {v_1, v_2,.....,v_N}$ and $\mathbf{A} \in \mathbb{R}^{N \times N}$. Note that $a_{ij}$ indicates whether node $v_i$ is connected to the node $v_j$. As $G$ is unweighted and undirected, $\mathbf{A} \in \{0,1\}^{N \times N}$ and $a_{ij}$ = $a_{ji}$. The degree of a vertex $v_i$ is given by $d_i = \sum_{j=1}^{N} a_{ij}$. We construct the degree matrix $\mathbf{D}$ which is a diagonal matrix with the degrees $d_1, d_2, \cdots , d_N$ of the respective vertices $v_1, v_2, \cdots , v_N$ as the diagonal elements. With the adjacency matrix $\mathbf{A}$ and the degree matrix $\mathbf{D}$, the graph Laplacian $\mathbf{L}$ is given by: 
\begin{equation}
    \mathbf{L} = \mathbf{D} - \mathbf{A}.
    \label{eq: Laplacian}
\end{equation}

Here, $\mathbf{L}$ is the unnormalized Laplacian matrix of the graph $G$ and $\mathbf{L} \in \mathbb{R}^{N \times N}$. 
The eigenvectors of $\mathbf{L}$ are calculated next. Let $\mathbf{L}$ has $N$ eigenvalues denoted by: $\lambda_1 \leq \lambda_2 \leq....\leq \lambda_N$ (\cite{von2007tutorial}). Further, let $\mathbf{u_1}, \mathbf{u_2}, \cdots , \mathbf{u_N}$ be the corresponding eigenvectors, which can be obtained by solving the generalized eigenproblem $\mathbf{L} \mathbf{u}_i$ = $\lambda_i \mathbf{D} \mathbf{u}_i$, where $i = \{1, 2, \cdots, N\}$. 

\subsection{Graph Spectral Clustering}
In this work, the moving objects are determined as the connected components (clusters) in the graph. Let us consider there are $k$ ($k\leq N$) moving objects which means there are $k$ clusters. The clusters are obtained by the spectral clustering algorithm following (\cite{shi2000normalized}):

\begin{itemize}[leftmargin=*]
    \itemsep0em
    \item Let $\mathbf{U}$ be the matrix whose columns are the first $k$ eigenvectors $\mathbf{u_1}, \mathbf{u_2},\cdots, \mathbf{u_k}$ where $\mathbf{U} \in \mathbb{R}^{N \times k}$.
    \item For $i= 1,\cdots , N$, let $\mathbf{y_i}$ be the vector corresponding to $i$th row of $\mathbf{U}$.
    \item Now the vectors ${\mathbf{y_i}}_{\{i=1,\cdots , N\}}$ are clustered using $k$-means algorithm into $k$ clusters $C_1, C_2,\cdots , C_k$. Note that $\mathbf{y_i} \in \mathbb{R}^{k}$. 
    \item This induces a set of clusters $A_1, A_2, \cdots , A_k$ on the original data, in which $e_j$ is assigned to $A_i$ if $y_j$ is assigned to $C_i$ (\cite{tung2010enabling}).
    
\end{itemize}

\begin{algorithm}[]
\KwIn{Sampled events in space-time (from the user).}
\KwOut{Moving objects as clusters.}

\nl  Construct a similarity graph from each event sample by using k-NN.

\nl Compute the adjacency matrix for the graph.

\nl Compute the unnormalized Laplacian.

\nl Compute the eigenvalues and eigenvectors.

\nl \SetKwFor{}{}{}{}\textbf{for} a range of values of $k$:

\hspace{\parindent}  Take the first $k$ eigenvectors and construct another matrix taking those eigenvectors as
columns.

\hspace{\parindent}   Take the first $k$ rows from that newly formed matrix.

\hspace{\parindent}   Obtain the clusters. 

\hspace{\parindent}   Compare and Update the maximum Silhouette Coefficient (SC) obtained so far.

\nl Get the moving objects as clusters and their corresponding labels for which SC is maximum.

    \caption{{\bf GSCEventMOD} \label{GSCEventMOD}}

\end{algorithm}

\subsection{Optimal Number of Clusters}
Determining the optimal number of clusters in data is a crucial task. Even though there are no foolproof methods available for determining its value, there are few widely used methods like the elbow method (\cite{syakur2018integration}), Rand Index (\cite{santos2009use}), adjusted Rand Index (\cite{hubert1985comparing}) and silhouette analysis (\cite{shutaywi2021silhouette}). In our work, we use the silhouette analysis strategy to find the actual number of clusters in the data, which, in our case, is the actual number of moving objects in a particular scene. With this strategy we do not need to have any prior knowledge about the exact number of clusters. A range of values for the number of clusters is inputted instead and the optimal value of the number of cluster is automatically determined. 

Silhouette value measures the difference between the within-cluster tightness and separation from the rest (\cite{rousseeuw_2002}). It is defined for each sample point and is composed of two scores:
\begin{itemize}[leftmargin=*]
    \itemsep0em
    \item ${a(i)}$: The mean distance between a sample point $i$ and all other points in the same cluster.
    \item ${b(i)}$: The mean distance between a sample point $i$ and all other points in all other clusters of which it is not a member.
\end{itemize}
Now the silhouette value $s(i)$ for that single sample is given as:
\begin{equation} \label{silhouette}
s(i) = \frac{b(i)-a(i)} {max(a(i),b(i))} ,
\end{equation}

Kauffman \textit{et al.} (\cite{kaufman_rousseeuw_1990}) proposed the term Silhouette Coefficient ${SC}$, which returns the maximum value of the mean $s(i)$ over all data points of a specific sequence. ${SC}$ is defined as:
\begin{equation}
    SC = \max_k  \Tilde{s}(k),
\end{equation}
where $\Tilde{s} (k)$ represents the mean $s(i)$ over the entire data of a specific sequence for a definite number of clusters $k$.   
The value of ${SC}$ is bounded in between $-1$ and $1$, where a score near to $1$ shows greater intra-cluster tightness and greater inter-cluster distance. A higher value means that the clusters are dense and well separated, leading to a meaningful representation of the clusters.

\begin{figure}
\begin{center}
\includegraphics[width = 0.6\textwidth]{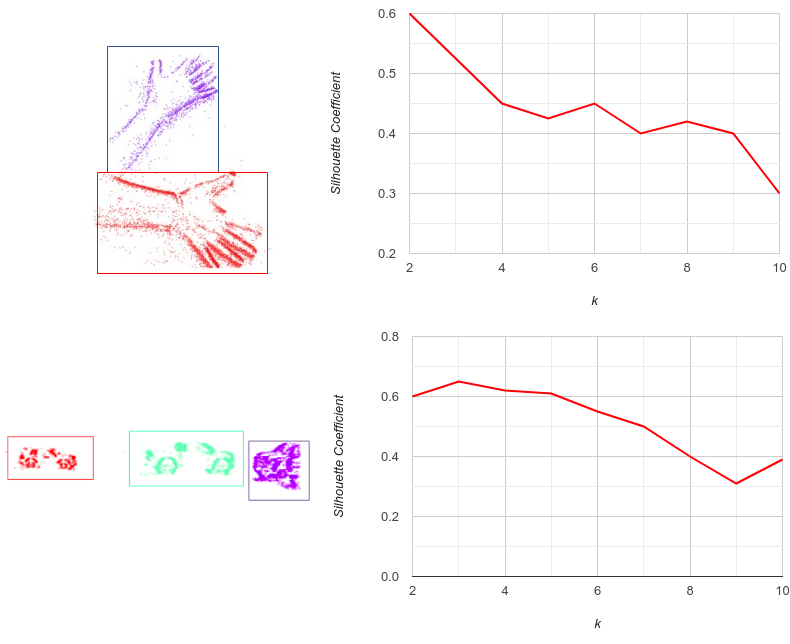}
\end{center}
   \caption{Few outputs from our framework and their corresponding Silhouette plots. Note how the maxima in the two plots are correctly representing the number of moving objects in the scenes (best viewed in color).}
    \label{fig:sample and sil plots}
\label{fig:sample and sil plots}
\end{figure}

In our work, we use ${SC}$ to determine the optimal value of $k$. We take some values of $k$, ranging from 2 to 10 (by the knowledge from the ground truth about the maximum number of moving objects) and plot their respective ${SC}$s. From those plots, we take the value of $k$ for which the magnitude of ${SC}$ is maximum. This $k$ is the required optimal number of clusters. This is how we get the actual number of moving objects in a scene. We show some sample $k$ vs $SC$ plots from our work in Fig. \ref{fig:sample and sil plots}. The whole methodology is summarized in Algorithm \ref{GSCEventMOD}.

\section{Experimental Results}
In this section, we evaluate our proposed method. First, we show the datasets used and metrics in Sections 4.1 and 4.2 respectively. In Section 4.2, we also compare our method against the state-of-the-art baselines (\cite{pikatkowska2012spatiotemporal,chen2018neuromorphic,hinz2017online}) and show how it is performing better than theirs.

We experimentally set the value of nearest neighbors at 30 for hands, 100 for cars and 25 for street sequences. 
All the experiments were performed on a computer with 2 Intel\textsuperscript{\tiny\textregistered} Xenon\textsuperscript{\tiny\textregistered} CPUs, each with a clock frequency of $3.50$ GHz.

\begin{figure*}
\begin{center}
\includegraphics[width = 0.81\textwidth]{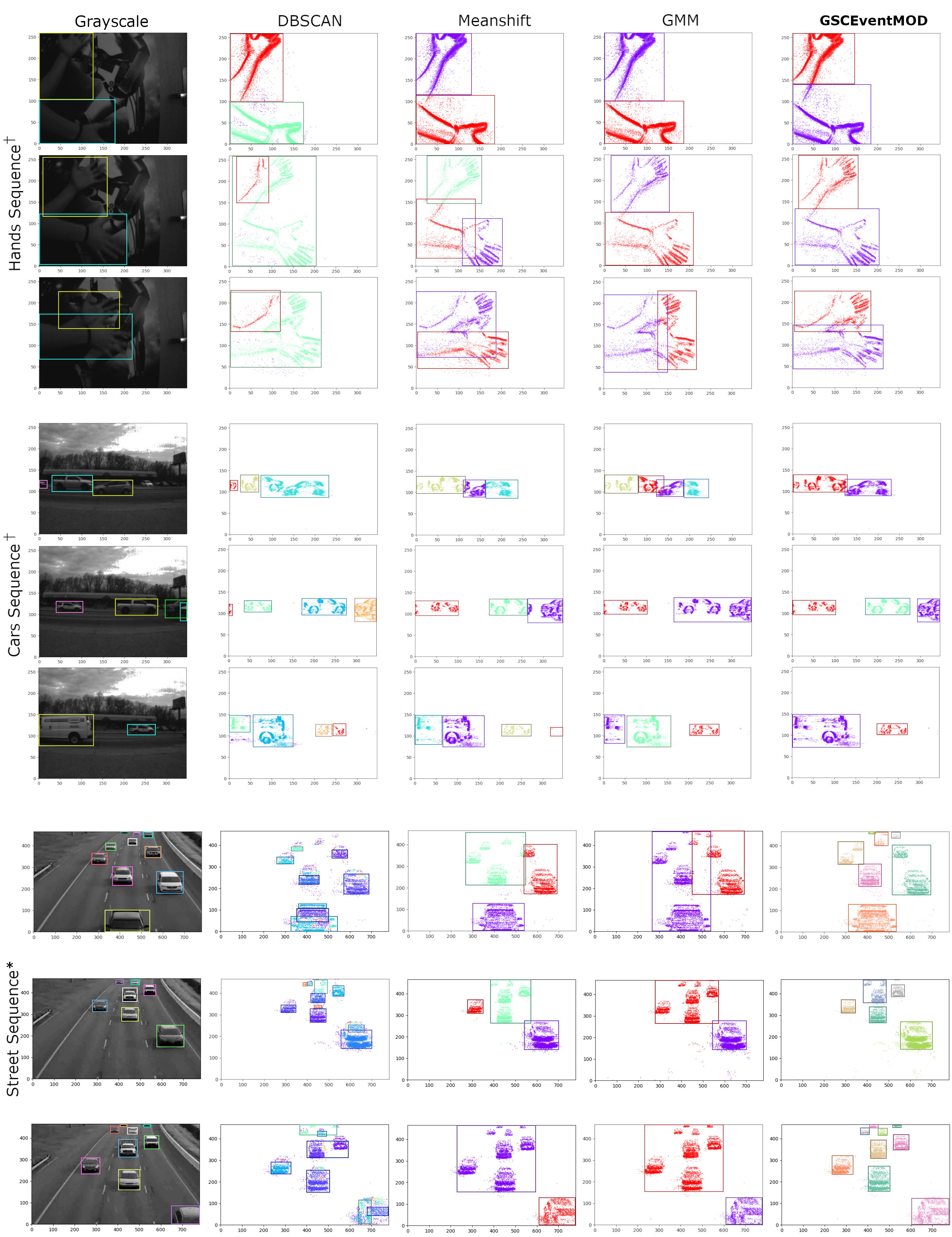}
\end{center}
   \caption{Visual results for GSCEventMOD and other SOTA methods on the DistSurf (\cite{almatrafi2020distance}) (marked with $\dagger$) and the synthetic dataset (marked with $*$). We show that our method performs significantly better than the other compared methods (best viewed in color).}
    \label{fig:comparison diagram}
\label{fig:comparison diagram}
\end{figure*}

\subsection{Datasets}

As the datasets used in the baseline methods (\cite{pikatkowska2012spatiotemporal,chen2018neuromorphic}) are not publicly available, we evaluate our algorithm on the DistSurf\footnote{\url{https://sites.google.com/a/udayton.edu/issl/software/dataset}} (\cite{almatrafi2020distance}) dataset, where the events are recorded using the IniVation DAViS346 camera. This camera has a $346 \times 260$ spatial resolution and outputs frames up to $60$ frames per second for RGB/ grayscale images and in microsecond resolution for the events. This dataset contains sequences (both events and their corresponding grayscale images) captured in both indoor and outdoor environments, involving multiple moving objects (cars, pedestrians, hands, etc.). Few sequences from this dataset (\cite{almatrafi2020distance}) are shown in Fig. \ref{fig:comparison diagram} (hands and cars). 

We further evaluate our method on a synthetic dataset generated using the v2e framework\footnote{\url{https://sites.google.com/view/video2events/home}} (\cite{hu2021v2e}). Using (\cite{hu2021v2e}), we convert the grayscale frames from a video to realistic events. The synthetic sequences are shown in Fig. \ref{fig:comparison diagram} (street).

\begin{table}[]
\centering
\caption{Evaluation metrics}
\label{tab:metric}
\begin{tabular}{|c|c|}
\hline
\textbf{Metrics} & \textbf{Formula} \\ \hline
True positive (TP) & ${\frac{\left|E \cap GT \right|}{\left|E \cup GT \right|} \ge 0.75}$ \\ \hline
False positive (FP) & ${\frac{\left|E \cap GT \right|}{\left|E \cup GT \right|} < 0.75}$ \\ \hline
False negative (FN) & No ground truth detected \\ \hline
Precision (P) & ${\frac{TP}{TP+FP}}$ \\ \hline
Recall & ${\frac{TP}{TP+FN}}$ \\ \hline
F measure & $\frac{2 \times P \times R} {P + R}$ \\ \hline
\end{tabular}
\end{table}

\begin{table*}[]
\centering
\caption{Comparison of GSCEventMOD with the state-of-the-art methods. Note that the precision and recall scores are in percentage. The best results are in bold.}
\label{tab:metric table}
\makebox[\linewidth]{
\scalebox{0.75}{
\begin{tabular}{c|ccc|ccc|ccc|ccc}
\hline
\multirow{2}{*}{Sequence} & \multicolumn{3}{c|}{{DBSCAN (\cite{chen2018neuromorphic})}} & \multicolumn{3}{c|}{{Meanshift (\cite{chen2018neuromorphic})}} & \multicolumn{3}{c|}{{GMM (\cite{pikatkowska2012spatiotemporal})}} & \multicolumn{3}{c}{\textbf{GSCEventMOD (Ours)}}       \\
                          & Recall   & Precision   & F measure  & Recall    & Precision    & F measure   & Recall  & Precision  & F measure & Recall & Precision & F measure \\ \hline
Hands (\cite{almatrafi2020distance})                     & 66.57    & 77.33       & 71.55      & 71.56     & 79.30        & 75.23       & 87.14   & 88.67      & 87.90     & \textbf{90.68}  & \textbf{91.56}     & \textbf{91.12}     \\ 
Cars (\cite{almatrafi2020distance})                      & 48.58    & 31.50       & 38.22      & 41.68     & 50.00           & 45.46       & 50.26   & 77.60      & 61.00     & \textbf{56.40}  & \textbf{82.29}     & \textbf{66.93}     \\ 
Street (\cite{hu2021v2e})                    & 44.94    & 34.87       & 39.27      & 38.95     & 52.00           & 44.53       & 43.78   & 67.59      & 53.14     & \textbf{58.64}  & \textbf{84.49}     & \textbf{69.23}     \\ \hline
\end{tabular}%
}
}
\end{table*}



\subsection{Evaluation Metrics and Comparison}

Multiple protocols are available for the detection of moving objects in frame-based cameras. We can apply many of them in event-based data too. As in (\cite{chen2018neuromorphic}), we accumulate the events corresponding to their grayscale frames in different time intervals. Now to decide how well the detected objects (in event-data) are located with respect to the ground truth (in frame-based data), we perform a coverage test over the whole dataset (\cite{pikatkowska2012spatiotemporal}) for the correctly detected objects $E$ (area of bounding box around moving objects in event-data) and the ground truth $GT$ (area of bounding box around moving objects in ground truth). As the Distsurf (\cite{almatrafi2020distance}) dataset does not have ground-truths for moving object detection, we have labeled them manually. The metrics used are shown in Table: \ref{tab:metric}. 

Visual comparisons of our method with the state-of-the-art approaches are shown in Fig. \ref{fig:comparison diagram} and the quantitative comparisons are shown in Table: \ref{tab:metric table}. From Fig. \ref{fig:comparison diagram}, we can see that the clusters represent the moving objects. However, we have noise surrounding those moving objects, which is generated mainly due to ambient disturbances and sensor defects. Despite that noise, our model could successfully detect the moving objects in comparison to other state-of-the-art methods (\cite{pikatkowska2012spatiotemporal,chen2018neuromorphic,hinz2017online}). We can also see that at the time of occlusions, while the other models fail to distinguish between the moving objects in the scene, our model is performing significantly better. Note that as the number of moving objects in a scene increases, the difference between the performance of our model and the previous methods becomes more prominent. 

\begin{figure*}
\begin{center}
\includegraphics[width = 0.9\textwidth]{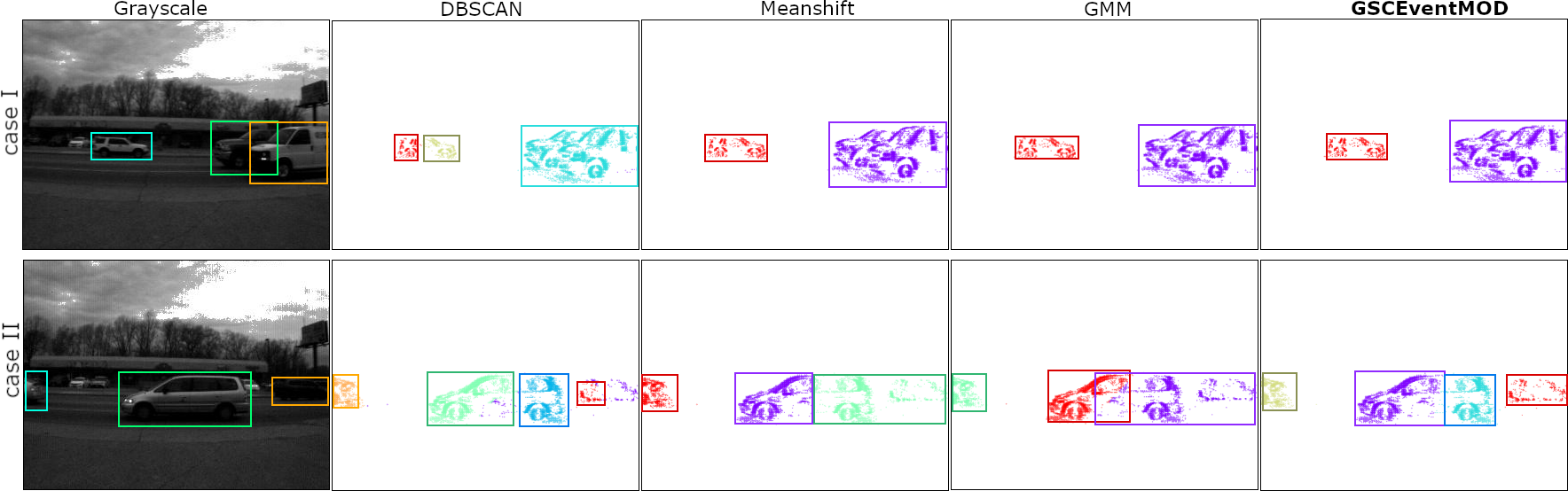}
\end{center}
   \caption{Some failure cases in GSCEventMOD and other SOTA methods. Case 1: When the moving objects are too close to each other.  Case 2: When one of the moving objects is significantly larger than all the others.}
   \label{fig:fail cases}
\end{figure*}

Nevertheless, there are some failure cases. For example, Fig. \ref{fig:fail cases} shows few samples, where the moving cars are so close to each other, that our model merges them into a single cluster (Case 1). Also, we could see that while a moving object is significantly larger than the others, the model is breaking it into multiple clusters (Case 2). However, even in such situations, our model performs better than the other methods, as can be seen from Fig. \ref{fig:fail cases}.

\section{Conclusions}
In this paper, we proposed a novel method, termed as GSCEventMOD for event-based moving object detection using graph spectral clustering. We demonstrated that detecting moving objects using neuromorphic vision sensors can perform well in the challenging situations like fast motions and abrupt changes in the lighting conditions. GSCEventMOD requires minimal pre-processing as scene dynamics is captured by the sensor itself. GSCEventMOD is shown to outperform some of the previous approaches in event-based vision. We validated GSCEventMOD with synthetic data and real-world data captured under varied environments to demonstrate its flexibility. Overall, we believe our method will be suitable for multiple computer vision applications like autonomous vehicles, robotics, remote surveillance, and others.

In future, we plan to incorporate suitable motion models. Another direction of future research would be to explore semi-supervised learning to improve the solution.

\bibliographystyle{unsrtnat}
\bibliography{references}  

\begin{thebibliography}{42}
\providecommand{\natexlab}[1]{#1}
\providecommand{\url}[1]{\texttt{#1}}
\expandafter\ifx\csname urlstyle\endcsname\relax
  \providecommand{\doi}[1]{doi: #1}\else
  \providecommand{\doi}{doi: \begingroup \urlstyle{rm}\Url}\fi

\bibitem[Chen et~al.(2020)Chen, Cao, Conradt, Tang, Rohrbein, and
  Knoll]{chen2020event}
Guang Chen, Hu~Cao, Jorg Conradt, Huajin Tang, Florian Rohrbein, and Alois
  Knoll.
\newblock Event-based neuromorphic vision for autonomous driving: a paradigm
  shift for bio-inspired visual sensing and perception.
\newblock \emph{IEEE Signal Processing Magazine}, 37\penalty0 (4):\penalty0
  34--49, 2020.

\bibitem[Lichtsteiner et~al.(2006)Lichtsteiner, Posch, and
  Delbruck]{lichtsteiner2006128}
Patrick Lichtsteiner, Christoph Posch, and Tobi Delbruck.
\newblock A 128 x 128 120db 30mw asynchronous vision sensor that responds to
  relative intensity change.
\newblock In \emph{IEEE International Solid State Circuits Conference-Digest of
  Technical Papers}, 2006.

\bibitem[Gallego et~al.(2020)Gallego, Delbruck, Orchard, Bartolozzi, Taba,
  Censi, Leutenegger, Davison, Conradt, Daniilidis, et~al.]{gallego2020event}
Guillermo Gallego, Tobi Delbruck, Garrick~Michael Orchard, Chiara Bartolozzi,
  Brian Taba, Andrea Censi, Stefan Leutenegger, Andrew Davison, Jorg Conradt,
  Kostas Daniilidis, et~al.
\newblock Event-based vision: A survey.
\newblock \emph{IEEE TPAMI}, 2020.

\bibitem[Kulchandani and Dangarwala(2015)]{kulchandani2015moving}
Jaya~S Kulchandani and Kruti~J Dangarwala.
\newblock Moving object detection: Review of recent research trends.
\newblock In \emph{IEEE International Conference on Pervasive Computing}, 2015.

\bibitem[Piccardi(2004)]{piccardi2004background}
Massimo Piccardi.
\newblock Background subtraction techniques: a review.
\newblock In \emph{IEEE International Conference on Systems, Man and
  Cybernetics}, 2004.

\bibitem[Agarwal et~al.(2016)Agarwal, Gupta, and Singh]{agarwal2016review}
Anshuman Agarwal, Shivam Gupta, and Dushyant~Kumar Singh.
\newblock Review of optical flow technique for moving object detection.
\newblock In \emph{IEEE International Conference on Contemporary Computing and
  Informatics}, 2016.

\bibitem[Bi et~al.(2020)Bi, Chadha, Abbas, Bourtsoulatze, and
  Andreopoulos]{bi2020graph}
Yin Bi, Aaron Chadha, Alhabib Abbas, Eirina Bourtsoulatze, and Yiannis
  Andreopoulos.
\newblock Graph-based spatio-temporal feature learning for neuromorphic vision
  sensing.
\newblock \emph{IEEE TIP}, 29:\penalty0 9084--9098, 2020.

\bibitem[Pi{\k{a}}tkowska et~al.(2012)Pi{\k{a}}tkowska, Belbachir, Schraml, and
  Gelautz]{pikatkowska2012spatiotemporal}
Ewa Pi{\k{a}}tkowska, Ahmed~Nabil Belbachir, Stephan Schraml, and Margrit
  Gelautz.
\newblock Spatiotemporal multiple persons tracking using dynamic vision sensor.
\newblock In \emph{IEEE CVPR Workshops}, 2012.

\bibitem[Mitrokhin et~al.(2018)Mitrokhin, Ferm{\"u}ller, Parameshwara, and
  Aloimonos]{mitrokhin2018event}
Anton Mitrokhin, Cornelia Ferm{\"u}ller, Chethan Parameshwara, and Yiannis
  Aloimonos.
\newblock Event-based moving object detection and tracking.
\newblock In \emph{IEEE/RSJ IROS}, 2018.

\bibitem[Stoffregen et~al.(2019)Stoffregen, Gallego, Drummond, Kleeman, and
  Scaramuzza]{stoffregen2019event}
Timo Stoffregen, Guillermo Gallego, Tom Drummond, Lindsay Kleeman, and Davide
  Scaramuzza.
\newblock Event-based motion segmentation by motion compensation.
\newblock In \emph{IEEE ICCV}, 2019.

\bibitem[Hinz et~al.(2017)Hinz, Chen, Aafaque, R{\"o}hrbein, Conradt, Bing, Qu,
  Stechele, and Knoll]{hinz2017online}
Gereon Hinz, Guang Chen, Muhammad Aafaque, Florian R{\"o}hrbein, J{\"o}rg
  Conradt, Zhenshan Bing, Zhongnan Qu, Walter Stechele, and Alois Knoll.
\newblock Online multi-object tracking-by-clustering for intelligent
  transportation system with neuromorphic vision sensor.
\newblock In \emph{Joint German/Austrian Conference on Artificial
  Intelligence}, 2017.

\bibitem[Chen et~al.(2018)]{chen2018neuromorphic}
Guang Chen et~al.
\newblock Neuromorphic vision based multivehicle detection and tracking for
  intelligent transportation system.
\newblock \emph{Journal of Advanced Transportation}, 2018.

\bibitem[Giraldo et~al.(2020)Giraldo, Javed, and Bouwmans]{giraldo2020graph}
Jhony~H. Giraldo, Sajid Javed, and Thierry Bouwmans.
\newblock Graph moving object segmentation.
\newblock \emph{IEEE TPAMI}, 2020.

\bibitem[Giraldo and Bouwmans(2021)]{giraldo2021graphbgs}
Jhony~H. Giraldo and Thierry Bouwmans.
\newblock {GraphBGS}: Background subtraction via recovery of graph signals.
\newblock In \emph{International Conference on Pattern Recognition}, 2021.

\bibitem[Xia et~al.(2021)Xia, Sun, Yu, Aziz, Wan, Pan, and Liu]{xia2021graph}
Feng Xia, Ke~Sun, Shuo Yu, Abdul Aziz, Liangtian Wan, Shirui Pan, and Huan Liu.
\newblock Graph learning: A survey.
\newblock \emph{IEEE Transactions on Artificial Intelligence}, 2021.

\bibitem[Ortega et~al.(2018)Ortega, Frossard, Kova{\v{c}}evi{\'c}, Moura, and
  Vandergheynst]{ortega2018graph}
Antonio Ortega, Pascal Frossard, Jelena Kova{\v{c}}evi{\'c}, Jos{\'e}~MF Moura,
  and Pierre Vandergheynst.
\newblock Graph signal processing: Overview, challenges, and applications.
\newblock \emph{Proceedings of the IEEE}, 106\penalty0 (5):\penalty0 808--828,
  2018.

\bibitem[Martin(2018)]{martin2018robust}
Lionel~J{\'e}r{\'e}mie Martin.
\newblock Robust and efficient data clustering with signal processing on
  graphs.
\newblock Technical report, EPFL, 2018.

\bibitem[Luo et~al.(2003)Luo, Wilson, and Hancock]{luo2003spectral}
Bin Luo, Richard~C Wilson, and Edwin~R Hancock.
\newblock Spectral clustering of graphs.
\newblock In \emph{International Workshop on Graph-Based Representations in
  Pattern Recognition}, 2003.

\bibitem[Von~Luxburg(2007)]{von2007tutorial}
Ulrike Von~Luxburg.
\newblock A tutorial on spectral clustering.
\newblock \emph{Statistics and computing}, 17\penalty0 (4):\penalty0 395--416,
  2007.

\bibitem[Ng et~al.(2001)Ng, Jordan, and Weiss]{ng2001spectral}
Andrew Ng, Michael Jordan, and Yair Weiss.
\newblock On spectral clustering: Analysis and an algorithm.
\newblock \emph{NeurIPS}, 14:\penalty0 849--856, 2001.

\bibitem[Panda et~al.(2017)Panda, Kuanar, and Chowdhury]{panda2017nystrom}
Rameswar Panda, Sanjay~K Kuanar, and Ananda~S Chowdhury.
\newblock Nystr{\"o}m approximated temporally constrained multisimilarity
  spectral clustering approach for movie scene detection.
\newblock \emph{IEEE Transactions on Cybernetics}, 48\penalty0 (3):\penalty0
  836--847, 2017.

\bibitem[Meila(2016)]{meila2016spectral}
Marina Meila.
\newblock Spectral clustering: a tutorial for the 2010’s.
\newblock \emph{Handbook of cluster analysis}, pages 1--23, 2016.

\bibitem[Shutaywi and Kachouie(2021)]{shutaywi2021silhouette}
Meshal Shutaywi and Nezamoddin~N Kachouie.
\newblock Silhouette analysis for performance evaluation in machine learning
  with applications to clustering.
\newblock \emph{Entropy}, 23\penalty0 (6):\penalty0 759, 2021.

\bibitem[Almatrafi et~al.(2020)Almatrafi, Baldwin, Aizawa, and
  Hirakawa]{almatrafi2020distance}
Mohammed Almatrafi, Raymond Baldwin, Kiyoharu Aizawa, and Keigo Hirakawa.
\newblock Distance surface for event-based optical flow.
\newblock \emph{IEEE TPAMI}, 42\penalty0 (7):\penalty0 1547--1556, 2020.

\bibitem[Menze and Geiger(2015)]{menze2015object}
Moritz Menze and Andreas Geiger.
\newblock Object scene flow for autonomous vehicles.
\newblock In \emph{IEEE CVPR}, 2015.

\bibitem[Huang et~al.(2019)Huang, Lin, Lin, Lee, and Chang]{huang2019deep}
Han-Yi Huang, Chih-Yang Lin, Wei-Yang Lin, Chien-Cheng Lee, and Chuan-Yu Chang.
\newblock Deep learning based moving object detection for video surveillance.
\newblock In \emph{IEEE International Conference on Consumer
  Electronics-Taiwan}, 2019.

\bibitem[Zhu et~al.(2020)Zhu, Yan, Tang, Chang, Li, and Yuan]{zhu2020moving}
Haidi Zhu, Xin Yan, Hongying Tang, Yuchao Chang, Baoqing Li, and Xiaobing Yuan.
\newblock Moving object detection with deep {CNNs}.
\newblock \emph{IEEE Access}, 8:\penalty0 29729--29741, 2020.

\bibitem[Rebecq et~al.(2021)Rebecq, Ranftl, Koltun, and
  Scaramuzza]{rebecq2019high}
Henri Rebecq, Ren{\'e} Ranftl, Vladlen Koltun, and Davide Scaramuzza.
\newblock High speed and high dynamic range video with an event camera.
\newblock \emph{IEEE TPAMI}, 43\penalty0 (6):\penalty0 1964--1980, 2021.

\bibitem[Reynolds(2009)]{reynolds2009gaussian}
Douglas~A Reynolds.
\newblock Gaussian mixture models.
\newblock \emph{Encyclopedia of biometrics}, 741:\penalty0 659--663, 2009.

\bibitem[Nguyen(2011)]{nguyen2011gaussian}
Thanh~Minh Nguyen.
\newblock \emph{Gaussian mixture model based spatial information concept for
  image segmentation}.
\newblock University of Windsor (Canada), 2011.

\bibitem[Khan et~al.(2014)Khan, Rehman, Aziz, Fong, and
  Sarasvady]{khan2014dbscan}
Kamran Khan, Saif~Ur Rehman, Kamran Aziz, Simon Fong, and Sababady Sarasvady.
\newblock {DBSCAN}: Past, present and future.
\newblock In \emph{International Conference on the Applications of Digital
  Information and Web technologies}, 2014.

\bibitem[Derpanis(2005)]{derpanis2005mean}
Konstantinos~G Derpanis.
\newblock Mean shift clustering.
\newblock \emph{Lecture Notes}, page~32, 2005.

\bibitem[Feng(2018)]{feng2018robust}
Xiaodong Feng.
\newblock Robust spectral clustering via sparse representation.
\newblock \emph{Recent Applications in Data Clustering}, page 155, 2018.

\bibitem[Zhou et~al.(2020)Zhou, Gallego, Lu, Liu, and Shen]{zhou2020event}
Yi~Zhou, Guillermo Gallego, Xiuyuan Lu, Siqi Liu, and Shaojie Shen.
\newblock Event-based motion segmentation with spatio-temporal graph cuts.
\newblock \emph{arXiv:2012.08730}, 2020.

\bibitem[Shi and Malik(2000)]{shi2000normalized}
Jianbo Shi and Jitendra Malik.
\newblock Normalized cuts and image segmentation.
\newblock \emph{IEEE TPAMI}, 22\penalty0 (8):\penalty0 888--905, 2000.

\bibitem[Tung et~al.(2010)Tung, Wong, and Clausi]{tung2010enabling}
Frederick Tung, Alexander Wong, and David~A Clausi.
\newblock Enabling scalable spectral clustering for image segmentation.
\newblock \emph{Pattern Recognition}, 43\penalty0 (12):\penalty0 4069--4076,
  2010.

\bibitem[Syakur et~al.(2018)Syakur, Khotimah, Rochman, and
  Satoto]{syakur2018integration}
MA~Syakur, BK~Khotimah, EMS Rochman, and Budi~Dwi Satoto.
\newblock Integration k-means clustering method and elbow method for
  identification of the best customer profile cluster.
\newblock In \emph{IOP Conference Series: Materials Science and Engineering},
  2018.

\bibitem[Santos and Embrechts(2009)]{santos2009use}
Jorge~M Santos and Mark Embrechts.
\newblock On the use of the adjusted rand index as a metric for evaluating
  supervised classification.
\newblock In \emph{International Conference on Artificial Neural Networks},
  2009.

\bibitem[Hubert and Arabie(1985)]{hubert1985comparing}
Lawrence Hubert and Phipps Arabie.
\newblock Comparing partitions.
\newblock \emph{Journal of classification}, 2\penalty0 (1):\penalty0 193--218,
  1985.

\bibitem[Rousseeuw(1987)]{rousseeuw_2002}
Peter~J. Rousseeuw.
\newblock Silhouettes: a graphical aid to the interpretation and validation of
  cluster analysis.
\newblock \emph{Journal of Computational and Applied Mathematics}, 20:\penalty0
  53--65, 1987.

\bibitem[Kaufman and Rousseeuw(1990)]{kaufman_rousseeuw_1990}
Leonard Kaufman and Peter~J. Rousseeuw.
\newblock \emph{Finding groups in data: an introduction to cluster analysis}.
\newblock Wiley, 1990.

\bibitem[Hu et~al.(2021)Hu, Liu, and Delbruck]{hu2021v2e}
Yuhuang Hu, Shih-Chii Liu, and Tobi Delbruck.
\newblock {v2e}: From video frames to realistic {DVS} events.
\newblock In \emph{IEEE CVPR Workshops}, 2021.

\end{thebibliography}






\end{document}